\documentclass{article}

\usepackage[utf8]{inputenc}
\usepackage{hyperref}
\usepackage{url}
\usepackage{booktabs}
\usepackage{amsfonts}
\usepackage{microtype}
\usepackage{subfigure}

\usepackage[final]{dai_2019}
\title{Iterative Update and Unified Representation for Multi-Agent Reinforcement Learning
}

\author{%
  Jiancheng Long \thanks{Equal contribution.}
  \\
  University of Science and Technology of China\\
  Hefei, China \\
  \texttt{sa517226@mail.ustc.edu.cn} \\
  \And
  Hongming Zhang $^*$
  \\
  Institute of Automation, Chinese Academy of Sciences \\
  Beijing, China \\
  \texttt{hongming.zhang@ia.ac.cn} \\
  \AND
  Tianyang Yu \\
  Nanchang University \\
  Nanchang, China \\
  \texttt{tianyang@email.ncu.edu.cn} \\
  \And
  Bo Xu \thanks{Corresponding author.} \\
  Institute of Automation, Chinese Academy of Sciences \\
  Beijing, China \\
  \texttt{boxu@ia.ac.cn} \\
}

\begin{document}

\maketitle

\begin{abstract}
Multi-agent systems have a wide range of applications in cooperative and competitive tasks. As the number of agents increases, nonstationarity gets more serious in multi-agent reinforcement learning (MARL), which brings great difficulties to the learning process. Besides, current mainstream algorithms configure each agent an independent network, so that the memory usage increases linearly with the number of agents which greatly slows down the interaction with the environment. Inspired by Generative Adversarial Networks (GAN), this paper proposes an iterative update method (IU) to stabilize the nonstationary environment. Further, we add first-person perspective and represent all agents by only one network which can change agents' policies from sequential compute to batch compute. Similar to continual lifelong learning, we realize the iterative update method in this unified representative network (IUUR). In this method, iterative update can greatly alleviate the nonstationarity of the environment, unified representation can speed up the interaction with environment and avoid the linear growth of memory usage. Besides, this method does not bother decentralized execution and distributed deployment. Experiments show that compared with MADDPG, our algorithm achieves state-of-the-art performance and saves wall-clock time by a large margin especially with more agents.
\end{abstract}

\section{Introduction}
A multi-agent system refers to a group of agents interact in a sharing environment, in which agents perceive the environment and form a policy conditioned on each other to accomplish a task\citep{vlassis2007concise} %\citep{weiss1999multiagent}\citep{horling2004survey}
. It is widely used in different fields, such as robotics\citep{stone2000multiagent},
distributed control\citep{weiss1999multiagent}, energy management\citep{riedmiller2000reinforcement}, etc. From the point of game theory\citep{nowe2012game}, %\citep{osborne2004introduction}\citep{bowling2000analysis}\citep{chalkiadakis2003multiagent}
these tasks can be divided into fully cooperative, fully competitive and mixed stochastic games. The complexity makes it difficult to design a fixed pattern to control the agents. A natural idea to solve it is learning on its own, which lead to the research on multi-agent reinforcement learning.

Multi-agent reinforcement learning (MARL)\citep{bucsoniu2010multi} %\citep{shoham2008multiagent}
is the combination of multi-agent system and reinforcement learning (RL)\citep{sutton2018reinforcement} %\citep{li2017deep}
. Agents interact with the common environment, perform actions and get rewards, learn a joint optimal policy by trial and error for multi-agent sequential decision making problems. In addition to the problems like sparse rewards and sample efficiency in reinforcement learning, multi-agent reinforcement learning encounters new problems such as the curse of dimensionality, nonstationary environment, multiple equilibria, etc. In addition, current mainstream algorithms tend to have a separate network structure for each agent, which undoubtedly poses a huge challenge to computing resources.

For the problem of nonstationary environment, the usual practice is to use centralized policy and global observation to transform the problem into a multi-agent problem with centralized control, but this method will encounter the curse of dimensionality and can not solve many tasks that require distributed deployment\citep{littman1994markov}\citep{hu2003nash} %\citep{bowling2002multiagent}
. An improved approach is to use centralized training to stabilize the environment and decentralized execution for distributed deployment\citep{peng2017multiagent}
%\citep{foerster2018counterfactual}
, which largely mitigates nonstationary environment. But usually all the agents in the system are learning simultaneously, which makes each agent actually face with a moving-target learning problem: its own optimal policy changes as other agents' policies change. Based on the idea of GAN\citep{goodfellow2014generative} %\citep{mirza2014conditional}\citep{chen2016infogan}
, we propose an iterative update method to stabilize the nonstationary environment. Divide the agents into the current learning agent and the agents waiting for learning, fix the strategies of agents in waiting list, only the current particular agent is trained, the problem is transformed into a single-agent case. Then each agent is switched regularly and gradually improved.

For the computing resources, current mainstream algorithms configure each agent an independent network. The memory usage increases linearly with the number of agents and the action of each agent needs to be computed separately by the network which also greatly slows down the interaction with environment. In this paper, all the agents are represented by one network, which greatly mitigates the demand for computing resources. The speed of interaction between the agent and the environment can also be accelerated by batch compute. To implement the iterative update strategy in this unified representation case, inspired by continual lifelong learning\citep{parisi2019continual}, we design a value fixing method based on Bellman Equation to achieve it. In this way, the strategy of the agents in waiting list will be fixed as much as possible.

We compare our method with MADDPG\citep{lowe2017multi}. Results show that our method both in fully-cooperative and mixed cooperative-competitive environments can effectively mitigate the nonstationarity and improve the performance. At the same time, the wall-clock time spent by the algorithm is greatly reduced. In particular, the advantages of the algorithm are more obvious when the number of agents increases.

The remainder of this paper is organized as follows. Section 2 reviews related work on multi-agent reinforcement learning. Section 3 presents the details of our method. The experimental setup and results are illustrated in Section 4. Section 5 and 6 give future research directions and summarize this paper.

% \begin{center}
%   \small
%   \url{http://www.adai.ai/call-for-papers.html}
% \end{center}

% The paper length is limited to 6 pages, with 1 additional page containing only
% bibliographic references. Authors may use as many pages of appendices (after
% the bibliography) as they wish, but reviewers are not required to read these.
% Any paper exceeding this length (except for the appendices) will automatically
% be rejected.

% \subsection{Style}
%
% Papers to be submitted to DAI 2019 must be prepared by \LaTeX{} using the style
% file \verb+dai_2019.sty+. As the review process is double blind, please use the
% style package as
% \begin{verbatim}
%   \usepackage{dai_2019}
% \end{verbatim}
% which automatically anonymizes the submission.
%
% The style file also provide an optional option \verb+final+, which creates a
% a camera-ready for your paper
% \begin{verbatim}
%   \usepackage[final]{dai_2019}
% \end{verbatim}
%
% The paper size, font sizes, margins, spaces between lines and headers or around
% figures/tables, etc, preset by the style file must {\em not} be modified. Any
% modification would lead to rejection without further notification.

\section{Related work}
Multi-agent reinforcement learning can be divided into centralized methods and decentralized methods according to specific tasks. We are concerned with methods that can extract decentralized policies for an agent to make decisions based on its own observation. In these tasks, centralized decisions are unattainable because global state and joint policy are unavailable.

This problem can be described as Partially Observable Markov Decision Process\citep{williams2007partially} %\citep{aviv2005partially}
(POMDP). Consider POMDP with $N$ agents $\left(\mathcal{S},\{\mathcal{O}\}_{i=1}^{N},\{\mathcal{A}\}_{i=1}^{N}, \mathcal{T},\{r\}_{i=1}^{N}\right). \mathcal{S} $
is a set global state, $\{\mathcal{O}\}_{i=1}^{N}$ is a set of observations for each agent, $\{\mathcal{A}\}_{i=1}^{N}$ is a set of actions, $\mathcal{T} : \mathcal{S} \times \mathcal{A}_{1} \times \dots \times \mathcal{A}_{N} \rightarrow \mathcal{S}$ denotes the state transition function.
 $r_{i} : \mathcal{S} \times  \mathcal{A}_{1} \times \ldots \times \mathcal{A}_{N} \rightarrow \mathbb{R}, i=1, \dots, N $
are the reward functions based on the joint actions. For each agent's policy
$r_{i} : \mathcal{S} \times \mathcal{A}_{1} \times \ldots \times \mathcal{A}_{N} \rightarrow \mathbb{R}, i=1, \dots, N$
, we have the joint policy $\pi :=\left(\pi_{\theta_{1}}, \ldots, \pi_{\theta_{N}}\right) \subset \Pi$. Because the rewards of the agents depend on the joint policy, we denote return for each agent as follows
$$R_{i}^{\pi}(\mathcal{S})=E\left(\sum_{k=0}^{\infty} \gamma^{k} r_{i, k+1} | s_{0} \in \mathcal{S}, \pi\right)$$
$\gamma$ is discount factor, $s_0$ is the initial state.

The goal is to get the equilibrium strategy to maximize each agent's return. We want to get the optimal policy $\pi^*$, for all $\pi_i$ we have
$$R_{i}^{\pi^{*}}(\mathcal{S}) \geq R_{i}^{\pi}(\mathcal{S}), \forall \pi_{i} \in \Pi_{i}, i=1, \dots, N$$

At this point, the problem can be considered as a stochastic game problem for multi-agents.
This definition is suitable for fully-cooperative, fully-competitive and mixed cooperative-competitive cases. Particularly, in fully-cooperative stochastic game, we have the same reward function for all agent, that is $r_{1}=\cdots=r_{N}$. The returns are also the same,
$R_{1}^{\pi}(\mathcal{S})=\cdots=R_{N}^{\pi}(\mathcal{S})$. For fully-competitive and mixed cooperative-competitive cases, the reward function for each agent depends on the given objective function of the environment.

For the good performance of Q-learning in single-agent cases, Tan\citep{tan1993multi} introduces independent Q-learning (IQL) for multi-agent reinforcement learning. This algorithm does not take the nonstationarity into consideration, each agent learns a $Q$ function independently. The independent approach hardly converge and it's no surprise that the performance is not as good as Q-learning\citep{watkins1992q} %\citep{matignon2012independent}\citep{mnih2013playing}\citep{mnih2015human}
in single-agent cases. %Gerald\citep{tesauro2004extending} proposes “Hyper-Q” Learning method to learn the action value function and estimates the strategies of other agents through Bayesian Inference. In this way, agents learn the Q function meanwhile consider other agents' policies and mitigate nonstationary problem to some extent.

Oliehoek\citep{oliehoek2008optimal} introduces a new paradigm of centralized training and decentralized execution and becomes the mainstream framework for decentralized tasks. This approach introduces global observed critics and decentralized actors for policy execution. The critic can minimize the estimation error of other agents and the actor can do decentralized decisions. MADDPG\citep{lowe2017multi} is the most popular algorithm in this paradigm. It extends DDPG\citep{lillicrap2015continuous} into multi-agent cases, each agent has a critic with global observation to direct the partial observed actor. MADDPG agents are able to perform coordination strategies in both cooperative, competitive and mixed environments.

VDN\citep{sunehag2017value} (value decomposition networks) establishes a link between centralized reinforcement learning and decentralized reinforcement learning. It decomposes the centralized Q function and learns each agent's Q value separately. Based on VDN, QMIX\citep{rashid2018qmix} designs a nonlinear function through network for value function decomposition. These methods can learn the accurate Q value for each agents which mitigates nonstationarity. However, such methods are only suitable for fully-cooperative environments.

COMA\citep{foerster2018counterfactual} combines the framework with counterfactual baseline. The baseline is obtained by fix other agents' actions which is a little similar with our methods. The difference is that we fixed the strategies of the agents rather than the specific actions. And COMA fixes other agents' actions to compute the baseline while our purpose is to learn the agent's policy.

In terms of computing resources, current mainstream algorithms simply skip the problem, but it will become more severe as the tasks' complexity increases. %Current mainstream algorithms tend to have a separate network structure for each agent, the memory usage increases linearly with the number of agents which undoubtedly poses a huge challenge to computing resources.
In some complex games like Pommerman\citep{peng2018continual}, Quake \uppercase\expandafter{\romannumeral 3}\citep{jaderberg2018human}, %soccer\citep{liu2019emergent}
and StarCraft %\citep{AlphaStar}
, a continuous league was created, and a large amount of competitors are trained by using parallel algorithm like population based training (PBT)
\citep{jaderberg2017population}, which undoubtedly poses a huge challenge to computing resources.

Focusing on the above problems, our method can greatly alleviate the nonstationarity and save computing resources.

\begin{figure*}[h]
  % \centering
  % \fbox{\rule[-.5cm]{0cm}{2cm} \rule[-.5cm]{5cm}{0cm}}
  \centering

  \subfigure[common method]{
  % \label{fig:subfig:a} %% 图的标签
  % \includegraphics[width=5 cm]{f1a.png}}
  \includegraphics[height=4 cm]{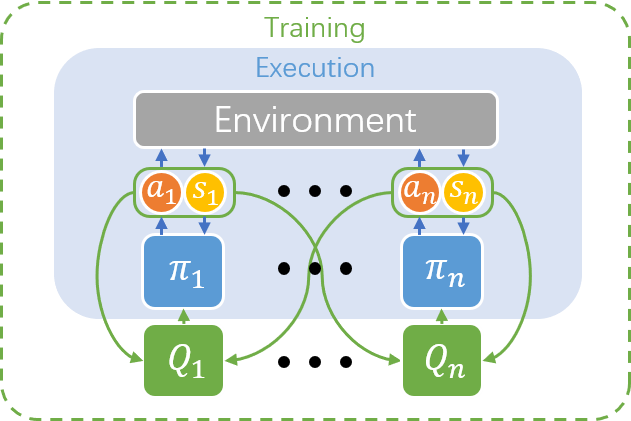}}
  \hspace{0.3cm}
  \subfigure[IU]{
  % \label{fig:subfig:b} %% 图的标签
  % \includegraphics[width=8 cm]{f1b.png}}
  \includegraphics[height=4 cm]{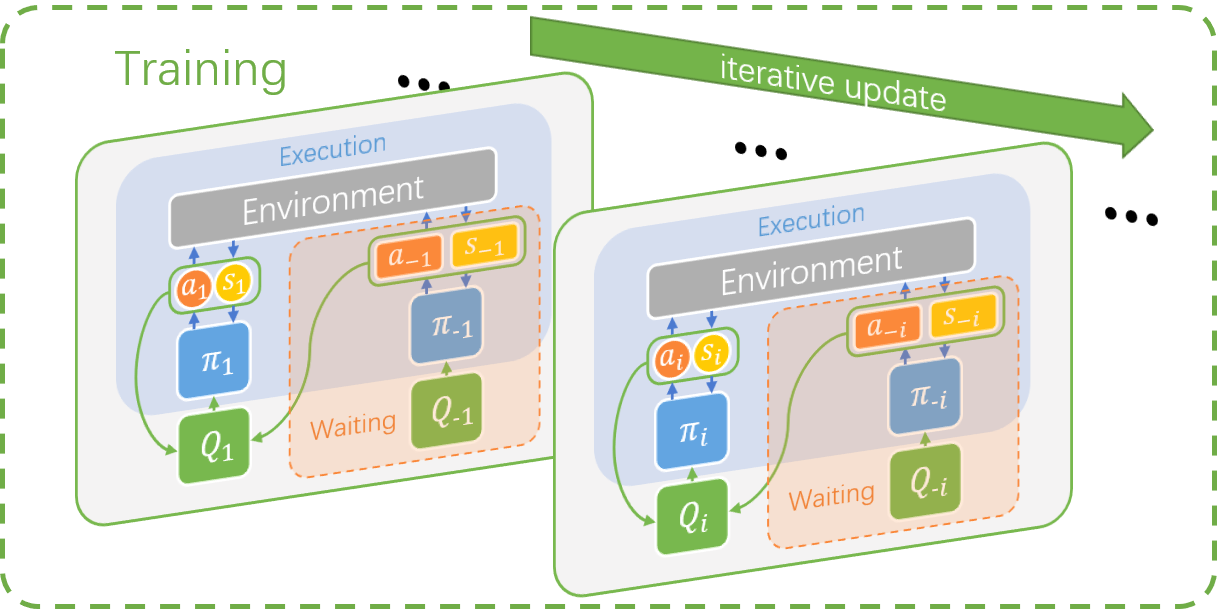}}

  \subfigure[IUUR]{
  % \label{fig:subfig:b} %% 图的标签
  % \includegraphics[width=0.8\textwidth]{f1c.png}}
  \includegraphics[height=4 cm]{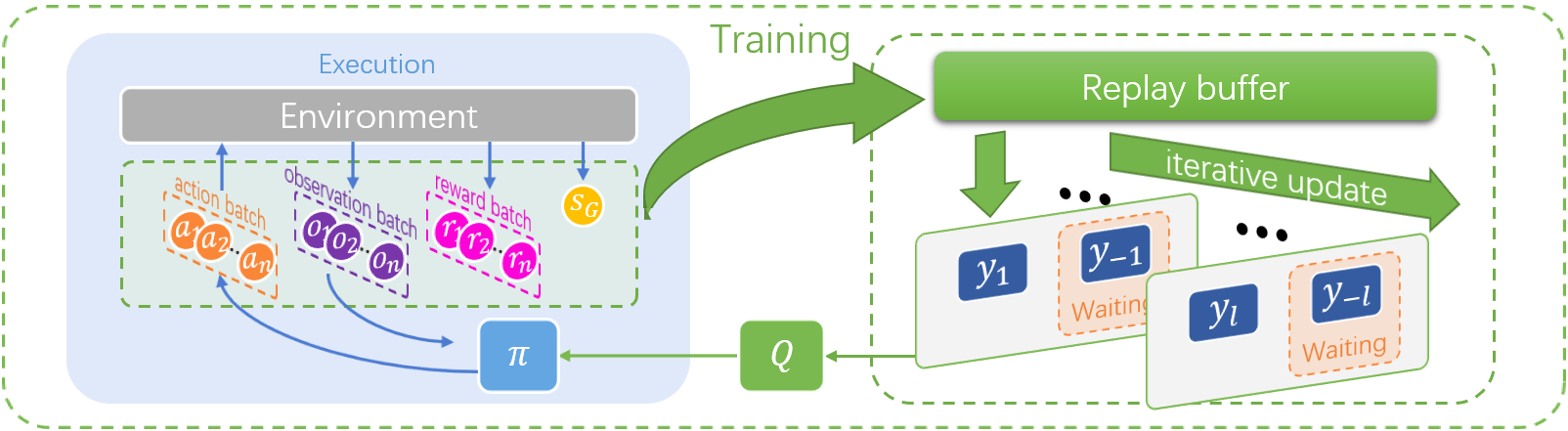}}

  \caption{(a) common method: all the agents are updated simultaneously (b) IU: the agents are updated iteratively (c) IUUR: represent all the agents in a unified network and update each agent iteratively}
  \label{fig1} %% 图的标签
\end{figure*}

\section{Method}
%Reinforcement learning methods in single-agent case are difficult to work in multi-agent environments. One of the main reasons is that each agent's strategy is constantly changing during the training process, this leads to an nonstationary environment\citep{jensen2005non}\citep{jia2000nonzero} for each agent because we have $P\left(s^{\prime} | s, a, \pi_{1}, \ldots, \pi_{N}\right) \neq P\left(s^{\prime} | s, a, \pi_{1}^{\prime}, \ldots, \pi_{N}^{\prime}\right), \forall \pi_{i} \neq \pi_{i}^{\prime}$. To some extent, it is meaningless for an agent to optimize its strategy based on this environment, especially when the number of agents increases and the environment is extremely nonstationary. Each agent faces with a moving-target learning problem\citep{vidal1998moving} and the optimal policy changes as the other agents' policies change.

A stable agent reduces the nonstationarity in the learning problem of the other agents which makes it easier to solve. We fix the policies of agents in waiting list and transform the problem into a single-agent case. Then we have $\pi_{j}=\pi_{j}^{\prime}, \forall j \in S_{\text {await}}, S_{\text {await}}$ is the waiting list and agents will be fixed. The critic has the global observation, the $Q$ action value function is $Q_{i} : \mathcal{S} \times \mathcal{A}_{1} \times \ldots \times \mathcal{A}_{N} \rightarrow \mathbb{R}, \mathcal{S}$
is the set of global state and $\mathcal{A}_{i}$ is agent $i$'s actions. For actor we have policy $\pi_{i} : \mathcal{O}_{i} \times \mathcal{A}_{i} \rightarrow[0,1], O_{i}$ is agent $i$'s perspective. When the agent $i$ is under training, let $\pi_{-i}$ denote the joint policy of all agents in $S_{\text {await}}$, Specifically, the $\pi_{-i}$ here is not changed.

Further, in order to improve computational efficiency and save wall-clock time. We represent all agents by only one network and add first-person perspective information to distinguish the roles of each agent. So we don't need to label or allocate separate parameters for each agent and they can make optimal polices based on its own perspective. Then the $Q$ action value function for critic is $Q_{i} : \mathcal{S} \times \mathcal{O}_{i} \times \mathcal{A}_{1} \times \ldots \times \mathcal{A}_{N} \rightarrow \mathbb{R}, \mathcal{O}_{i}$ is agent $i$'s perspective. For each actor, the observation is unchanged, so the policy remains
$\pi_{i} : \mathcal{O}_{i} \times \mathcal{A}_{i} \rightarrow[0,1]$. Here we use deterministic policy $\mu_{i} : \mathcal{O}_{i} \rightarrow \mathcal{A}_{i}$ like MADDPG. In this way, the memory usage won't increase regardless how many agents in the environment and the actions output from networks can be transformed from sequential compute to batch compute.

A critical problem is how to realize the iterative update method in this unified representative network. If all agents share a common network, it's no easy to fix the agents in waiting list. When we update the parameters in order to update one agent's policy, the others' will also change. Then the iterative update method fails and that's what we want to avoid.

We consider the problem as continual lifelong learning\citep{parisi2019continual} %\citep{thrun2012explanation}\citep{koper2004new}
, there are usually three types of methods. a) retraining with regularization to prevent catastrophic forgetting, b) extending network to represent new tasks, c) selective retraining with possible expansion. Retraining with regularization is not suitable for reinforcement learning because the policy update conflicts with the regularization which will impede the improvement of policy. Network expansion is obviously opposed to our purpose of saving memory usage. Still, inspired by the lifelong learning's thought, we focus on Bellman Equation\citep{peng1992generalized} and propose a value fixing method to achieve it. For current learning agent, Q value is computed by Bellman Equation, for agent in waiting list, Q value output from critic directly. In concrete, we denote experience replay buffer $\mathcal{D}$ includes global state, observations, rewards as follows:
$$
\left(s, o_{1}, \ldots, o_{N}, a_{1}, \ldots, a_{N}, r_{1}, \ldots, r_{N}, s^{\prime}, o_{1}^{\prime}, \ldots, o_{N}^{\prime}\right)
$$
For the current learning agent i, the Q value is updated follows Bellman Equation
$$
y_{i}=r_{i}+\gamma Q^{\prime}\left.\left(s, o_{i}, a_{1}^{\prime}, \ldots, a_{N}^{\prime}\right)\right|_{a_{j}^{\prime}=\mu^{\prime}\left(o_{j}\right)}, j=1, \ldots, N
$$
Let $-i$ denote other agents except $i$, we have $-i \in S_{\text { await }}$. Q value is directly output from the network
$$
y_{-i}=Q^{\prime}\left.\left(s, o_{-i}, a_{1}^{\prime}, \ldots, a_{N}^{\prime}\right)\right|_{a_{j}^{\prime}=\mu^{\prime}\left(o_{j}\right)}, j=1, \ldots, N
$$
Where $\mu^{\prime}$ is the target policy network.
The action-value function $Q$ is updated as:
$$
\mathcal{L}=E\left(Q\left(s, o_{i}, a_{1}, \ldots, a_{N}\right)-y\right)^{2}
$$
The samples $y$ from both $y_i$ and $y_{-i}$.
For the deterministic policy $\mu$, updated by gradient ascent as:
$$
\nabla_{\theta} \mathcal{J}\left(\mu | o_{i}\right)=E\left(\nabla_{\theta} \mu\left(o_{i} | \theta\right) \nabla_{a_{i}} Q\left.\left(s, o_{i}, a_{1}, \ldots, a_{N}\right)\right|_{a_{i}=\mu\left(o_{i}\right)}\right)
$$
In theory, when other agents' strategies are optimal with the current Q function, we have
{\scriptsize $$\nabla_{\theta} \mathcal{J}\left(\mu | o_{-i}\right)=E\left(\nabla_{\theta} \mu\left(o_{-i} | \theta\right) \nabla_{a_{-i}} Q\left.\left(s, o_{-i}, a_{1}, \ldots, a_{N}\right)\right|_{a_{-i}=\mu\left(o_{-i}\right)}\right)=\mathbf{0} $$}
Which means their policies will not change. But in practice, the Q function is represented with a neural network which is a nonlinear function. Consider the limited samples and the gradient ascent\citep{boyd2004convex} method's error, it is impossible to hold the policy gradient\citep{sutton2000policy} exactly equal to zero. Luckily we can usually guarantee a smaller $L_2$ norm of its gradient as follows:
$$
\left \| \nabla_{\theta} \mathcal{J}\left(\mu | o_{-i}\right)\right\|^{2}<\left\|\nabla_{\theta}^{\prime} \mathcal{J}\left(\mu | o_{-i}\right)\right\|^{2}
$$
Here $\nabla_{\theta}^{\prime}$ uses Bellman Equation also for agents $-i \in S_{\text {await}}$, the update is the same as:
$$
y_{-i}= r_{-i}+\gamma Q^{\prime}\left.\left(s, o_{-i}, a_{1}^{\prime}, \ldots, a_{N}^{\prime}\right)\right|_{a_{j}^{\prime}=\mu^{\prime}\left(o_{j}\right)}, j=1, \ldots, N
$$
Though we can't hold the equation accurately, this method can substantially fix the other agents' polices which mitigate the nonstationarity effectively in practice. The pseudo code of the algorithm is in $\bf Algorithm\ \ref{alg1}$.

% \begin{algorithm}
%   Initialize critic $Q\left(s, o_{i}, a_{1}, \ldots, a_{N} | \theta^{Q}\right)$ and actor $\mu\left(o_{i} | \theta^{\mu}\right)$ with random weights $\theta^Q$ and $\theta^{\mu}$
% \End{algorithm}

\begin{algorithm}[h]
\caption{IUUR}
\label{alg1}
\begin{algorithmic}[0]
  \STATE Initialize critic $Q\left(s, o_{i}, a_{1}, \ldots, a_{N} | \theta^{Q}\right)$ and actor $\mu\left(o_{i} | \theta^{\mu}\right)$ with random weights $\theta^Q$ and $\theta^{\mu}$
  \STATE Initialize target network $Q^{\prime}$ and $\mu^{\prime}$ with weights $\theta^{Q^{\prime}} \leftarrow \theta^{Q}, \theta^{\mu^{\prime}} \leftarrow \theta^{\mu}$
  \STATE Initialize replay buffer $\mathcal{D}$
  \STATE Learning agent $l=1$
  \FOR{$episode=1\ to\ M$}
    \STATE Initialize a random process $\mathcal{N}$ for action exploration
    \STATE Receive initial state $s$ and observation $o_i,i=1,\ldots,N$
    \IF{$episode\ mod\ K=0$}
      \STATE $l \leftarrow l+1$
      \ENDIF
    \FOR{$t=1, T$}
      \STATE For each agent $i$, select action $a_{i}=\mu\left(o_{i}\right)+\mathcal{N}_{t}$
      \STATE Execute actions $\vec {\boldsymbol a}=\left(a_{1}, \ldots, a_{N}\right)$
      \STATE Get reward $r_i$, new state $s^{\prime}$ and observation $o_{i}^{\prime}, i=1, \ldots, N$
      % \STATE Store$\left(s, o_{1}, \ldots, o_{N}, a_{1}, \ldots, a_{N}, r_{1}, \ldots, r_{N}, s^{\prime}, o_{1}^{\prime}, \ldots, o_{N}^{\prime}\right)$ in replay buffer $\mathcal{D}$
      \STATE Store$\left(s, { \vec {\boldsymbol o}, \vec {\boldsymbol a}, \vec {\boldsymbol r},} s^{\prime}, {\vec {\boldsymbol o^{\prime}}}\right)$ in replay buffer $\mathcal{D}$
      \STATE $s \leftarrow s^{\prime}$
      \STATE Sample a batch of $B$ transitions from $\mathcal{D}$
      \STATE \centering $\left(s^{j}, \vec {\boldsymbol o^{j}}, \vec {\boldsymbol a^{j}}, \vec {\boldsymbol r^{j}}, s^{\prime j}, \vec {\boldsymbol o^{\prime j}}\right), j=1, \ldots, B$
      \STATE \flushleft Set $y_{l}=r_{l}+\gamma Q^{\prime}\left.\left(s, o_{l}, {\vec {\boldsymbol a^{\prime}}}\right)\right|_{a_{i}^{\prime}=\mu^{\prime}\left(o_{i}\right)}, i=1, \ldots, N$
      \STATE Set $y_{-l}=Q^{\prime}\left.\left(s, o_{-l}, {\vec {\boldsymbol a^{\prime}}}\right)\right|_{a_{i}^{\prime}=\mu^{\prime}\left(o_{i}\right)}, i=1, \ldots, N$
      \STATE \flushleft Update critic by minimizing the loss
      \STATE \center $\mathcal{L}_{\theta^{Q}}=\frac{1}{N B} \sum_{i} \sum_{j}\left(y^{j}-Q\left(s^{j}, o_{i}^{j}, \vec {\boldsymbol a^{j}}\right)\right)^{2}$
      \STATE \flushleft Update the actor policy using the sampled policy gradient
      \STATE {\scriptsize \centering $\nabla_{\theta} \mathcal{J}\left(\mu | o_{i}\right) \approx \frac{1}{N B} \sum_{i} \sum_{j}\left(\nabla_{\theta^{\mu}} \mu\left(o_{i}\right) \nabla_{a_{i}} Q\left.\left(s^{j}, o_{i}^{j}, \vec {\boldsymbol a^{j}}\right)\right|_{a_{i}=\mu\left(o_{i}\right)}\right)$}
      \STATE \flushleft Update target network parameters
      \STATE \centering $\theta^{\mu^{\prime}} \leftarrow \tau \theta^{\mu}+(1-\tau) \theta^{\mu^{\prime}}$
    \ENDFOR
  \ENDFOR
\end{algorithmic}
\end{algorithm}

\section{Experiments and Results}
We evaluate the algorithm in fully-cooperative and mixed cooperative-competitive environments. In order to compare the influence of the number of agents on the algorithm, we designed the control groups by increasing the number of agents. Besides, all environments' state space and action space are continuous which are designed as follows.

\subsection{Environments}

\subsubsection{Fully-cooperative environments: Spread}
Agents perceive the environment on its own perspective and cooperate with each other to reach different destinations (the black points). In this environment, if agents collide during the movement, the agents will be punished. That is to say the agents must learn to reach the specified destinations without colliding with other agents. We set up a simple environment with three agents (Spread\_3) and a complex environment with ten (Spread\_10)as shown in figure \ref{fig2}:

\begin{figure}[h]
  % \centering
  % \fbox{\rule[-.5cm]{0cm}{2cm} \rule[-.5cm]{5cm}{0cm}}
  \centering

  \subfigure[Spread\_3]{
  % \label{fig:subfig:a} %% 图的标签
  \fbox{\includegraphics[width=3 cm]{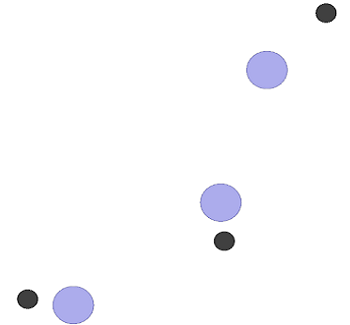}}}
  \hspace{0.5cm}
  \subfigure[Spread\_10]{
  % \label{fig:subfig:b} %% 图的标签
  \fbox{\includegraphics[width=3 cm]{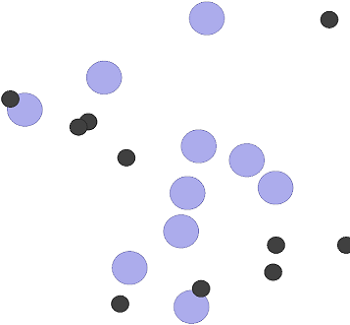}}}

  \caption{Spread environments. Agents should reach different destinations (the black points) without colliding with each other.}
  \label{fig2} %% 图的标签
\end{figure}

\subsubsection{Mixed cooperative-competitive environments: Predator-Prey}
\begin{figure}[h]
  % \centering
  % \fbox{\rule[-.5cm]{0cm}{2cm} \rule[-.5cm]{5cm}{0cm}}
  \centering

  \subfigure[Predator\_3-Prey\_1]{
  % \label{fig:subfig:a} %% 图的标签
  \fbox{\includegraphics[width=3 cm]{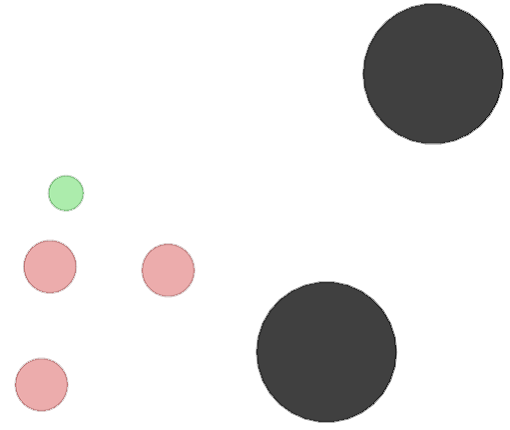}}}
  \hspace{0.5cm}
  \subfigure[Predator\_6-Prey\_2]{
  % \label{fig:subfig:b} %% 图的标签
  \fbox{\includegraphics[width=3 cm]{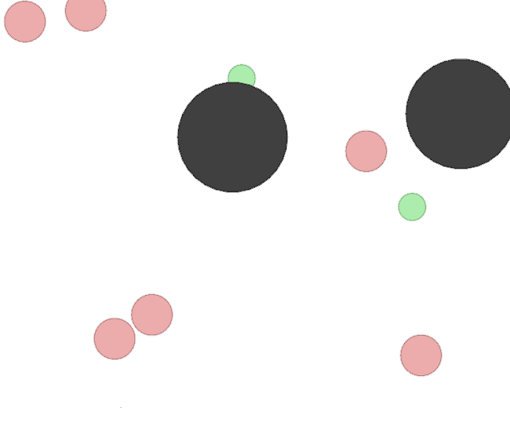}}}

  \caption{Predator-Prey environments. Any of the predators collides with the prey will get better return and win the game. Otherwise the prey will escape successfully and get better return.}
  \label{fig3} %% 图的标签
\end{figure}

Agents are divided into predators and preys. The predators need to cooperate with each other to catch the preys. The prey needs to find a way to escape as much as possible. Two obstacles (black circle) will render randomly to block the way. If any of the predators collides with the prey, predators win and otherwise the prey wins. We set up three chase one as simple scenes and six chase two as complex scenes, which are respectively recorded as Predator\_3-prey\_1 and Predator\_6-prey\_2, as shown in figure \ref{fig3}

\subsection{Results}
We compare IU (Iterative Update) and IUUR (Iterative Update and Unified Representation) on the basis of MADDPG. IU only uses iterative update method and IUUR adopts both iterative update and unified representation. The network structure is consistent with MADDPG, a two-layer ReLU MLP with 64 units per layer. Using the Adam optimizer, we set the soft update parameter $\tau=0.01$ and train each model 100,000 episodes. We only fine-tuned the new hyperparameters $K$ to control the frequency of iterative update. In our experiments, we set $K=5000$. Experiments show that our algorithm not only get good performance, but also improve computational efficiency.The source code of our algorithm implementation is available online
( \small
  \url{https://github.com/DreamChaser128/IUUR-for-Multi-Agent-Reinforcement-Learning}).

\subsubsection{Performance}
We run five random seeds for each environment and compare the performance among MADDPG, IU and IUUR.

\begin{figure}[h]
  % \centering
  % \fbox{\rule[-.5cm]{0cm}{2cm} \rule[-.5cm]{5cm}{0cm}}
  \centering

  \subfigure[Spread\_3]{
  % \label{fig:subfig:a} %% 图的标签
  % \fbox{\includegraphics[width=3 cm]{f4a.png}}}
  \includegraphics[width=3.9 cm]{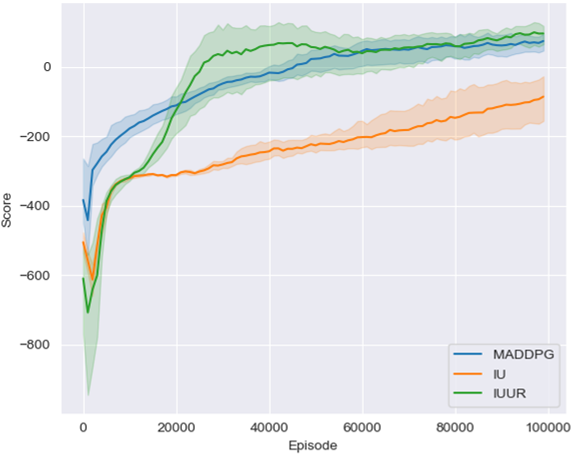}}
  \hspace{0.1cm}
  \subfigure[Spread\_10]{
  % \label{fig:subfig:b} %% 图的标签
  % \fbox{\includegraphics[width=3 cm]{f4b.png}}}
  \includegraphics[width=3.9 cm]{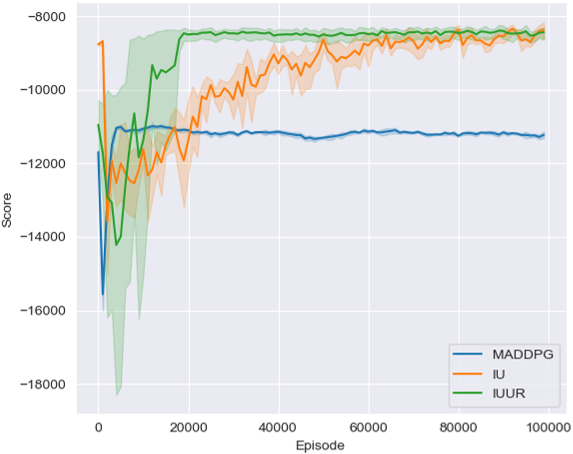}}

  \caption{Agents reward curves on fully-cooperative environments Spread\_3 and Spread\_10.}
  \label{fig4} %% 图的标签
\end{figure}

The results of fully-cooperative environments (Spread\_3 and Spread\_10) show in figure \ref{fig4}. In Spread\_3, we can find the IUUR converges quickly and after 20,000 episodes, it has exceeded MADDPG and maintained a steady rise. It's surprising that IU is inferior to MADDPG though it still has an upward trend after the maximum training steps. This may show that the nonstationarity of the environment in simple environment is not particularly serious and the new hyperparameters $K$ may impede the learning efficiency of agents in waiting list. In Spread\_10, it can be clearly seen that IUUR and IU both greatly exceed the performance of MADDPG, which indicates that as the number of agents increases, the nonstationarity gets worse, and iterative update can effectively alleviate the problem.

\begin{figure}[h]
  % \centering
  % \fbox{\rule[-.5cm]{0cm}{2cm} \rule[-.5cm]{5cm}{0cm}}
  \centering

  \subfigure[predator comparison]{
  % \fbox{\includegraphics[width=3 cm]{f4a.png}}}
  \includegraphics[width=3.9 cm]{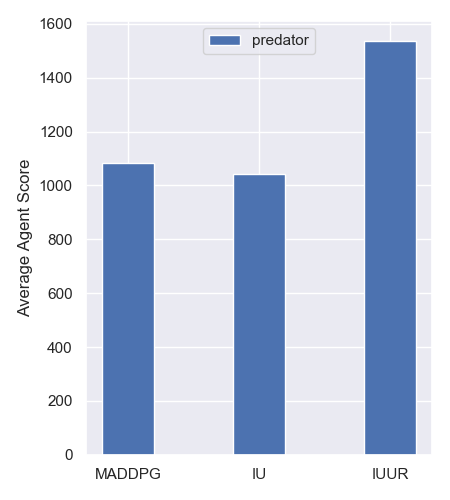}}
  \hspace{0.1cm}
  % comparison
  \subfigure[prey comparison]{
  % \fbox{\includegraphics[width=3 cm]{f4b.png}}}
  \includegraphics[width=3.9 cm]{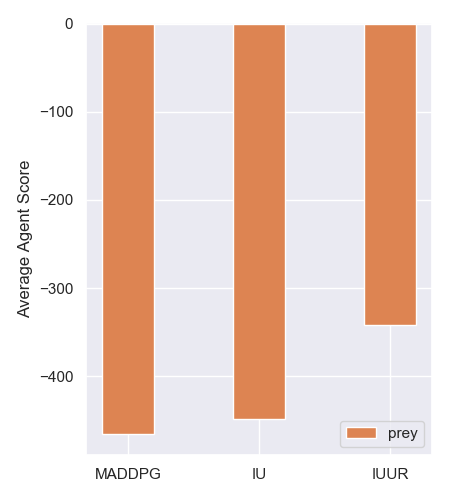}}

  \caption{Performance comparison in Predator\_3-Prey\_1. In (a), predators are replaced by IU and IUUR, we can find that IUUR outperforms MADDPG a lot and IU's performance is slightly worse than that of MADDPG. In (b), both IU and IUUR get a better reward than MADDPG.}
  \label{fig5} %% 图的标签
\end{figure}

For the mixed cooperative-competitive environments Predator-Prey. We set MADDPG vs MADDPG as baseline, then replace the predators or preys with our IU and IUUR to compete with MADDPG. By comparing the reward of our algorithms with the MADDPG agents, we can clearly get the performance of each methods.

Environment Predator\_3-Prey\_1 shows in figure \ref{fig5}, when we replace the predators, IUUR outperforms MADDPG a lot. IU's performance is slightly worse than that of MADDPG which is out of our expectation. In theory, we believe that IU can fix the agents' strategies in waiting list, IUUR can only guarantee a smaller $L_2$ norm of its gradients which can only alleviate the nonstationarity to some extent. In this sense, we think IU will be better than IUUR, but the experimental result shows opposite performance. The reason may be the new hyperparameters $K$, though IU can stable the environment, it will impede the learning efficiency of agents in waiting list at the same time which is very similar to environment Spread\_3. When we replace the prey, both IU and IUUR can get a better reward and IUUR owns the highest reward.

\begin{figure}[h]
  % \centering
  % \fbox{\rule[-.5cm]{0cm}{2cm} \rule[-.5cm]{5cm}{0cm}}
  \centering

  \subfigure[predator comparison]{
  % \fbox{\includegraphics[width=3 cm]{f4a.png}}}
  \includegraphics[width=3.9 cm]{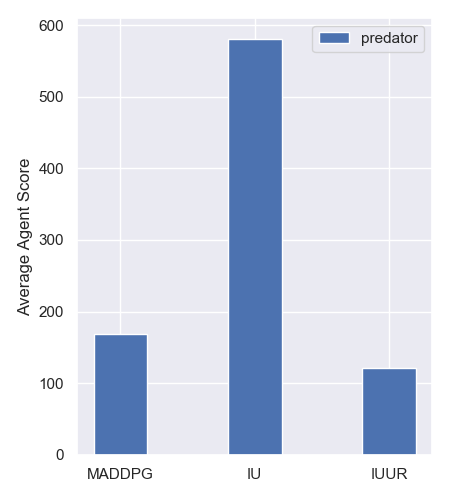}}
  \hspace{0.1cm}
  % comparison
  \subfigure[prey comparison]{
  % \fbox{\includegraphics[width=3 cm]{f4b.png}}}
  \includegraphics[width=3.9 cm]{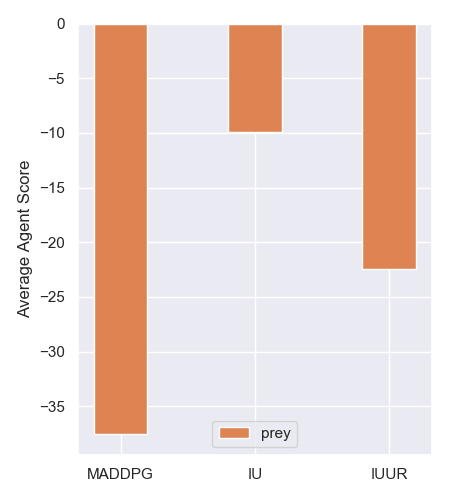}}

  \caption{Performance comparison in Predator\_6-Prey\_2. In (a), predators are replaced by IU and IUUR, we can find that IU outperforms MADDPG a lot and IU's performance is worse than that of MADDPG. In (b), IU and IUUR perform much better than MADDPG}
  \label{fig6} %% 图的标签
\end{figure}

Environment Predator\_6-Prey\_2 shows in figure \ref{fig6}, when we replace the predators, IU outperforms MADDPG a lot. IUUR's performance is worse than that of MADDPG. The reason is that as the number of agents increases, nonstationarity arises in multi-agent reinforcement learning gets more serious. For IU, each agent has its own network and iterative update can alleviate the nonstationarity effectively. But for IUUR, though iterative update can alleviate the nonstationarity, the update error is introduced by unified representation which leads to the inaccuracy of $Q$ value and poor performance. When we replace the preys, both IU and IUUR get a better reward and IU owns the highest reward. Overall, although there is a slight gap between IUUR and MADDPG, considering that IUUR can learn faster and use less memory space, so even if there is a slight difference in performance, it is still within acceptable range. Moreover, we only simply control the learning frequency of iterative update hyperparameter $K$ through experience, which plays a key role on the performance and can be improved in the future work.

\subsubsection{Computational efficiency}
We compared the differences in training time and the speed of interaction with environment. Table \ref{t1} and table \ref{t2} show the details in different environments. All results were generated on 2.20GHz Intel Xeon E5-2630 and 2 GeFore GTX 1080Ti graphics cards based machine running ubuntu.

Obviously, IUUR saves a lot of time in both training and interaction by a large margin especially with more agents. That is to say, IUUR can be used in large scale multi-agent reinforcement learning without the linear growth of wall-clock time.

\begin{table}[h]
  \caption{Computational efficiency in fully-cooperative environments.}
  \label{t1}
  \centering
  \begin{tabular}{ccccccc}
    \toprule
    environment & \multicolumn{3}{c}{spread\_3} \\
    algorithm & Baseline & IU & IUUR \\
    \midrule
    training\_time(h) & 5.76 & 4.25 & \bf3.47 \\
    Interaction\_time(s) & 0.0011 & 0.0011 & \bf0.0007 \\
    \bottomrule
  \end{tabular}
  \\
  \begin{tabular}{ccccccc}
    \toprule
    environment & \multicolumn{3}{c}{spread\_10}\\
    algorithm & Baseline & IU & IUUR \\
    \midrule
    training\_time(h) & 34 & 31 & \bf28\\
    Interaction\_time(s) & 0.0037 & 0.0036 & \bf0.0012\\
    \bottomrule
  \end{tabular}
\end{table}

\begin{table}[h]
  \caption{Computational efficiency in mixed cooperative-competitive environments.}
  \label{t2}
  \centering
  \begin{tabular}{ccccccc}
    \toprule
    environment & \multicolumn{3}{c}{Predator\_3-prey\_1} \\
    algorithm & Baseline & IU & IUUR \\
    \midrule
    training\_time(h) & 6.4 & 3.6 & \bf1.7 \\
    Interaction\_time(s) & 0.0014 & 0.0015 & \bf0.001 \\
    \bottomrule
  \end{tabular}
  \\
  \begin{tabular}{ccccccc}
    \toprule
    environment & \multicolumn{3}{c}{Predator\_6-prey\_2}\\
    algorithm & Baseline & IU & IUUR \\
    \midrule
    training\_time(h) & 8.7 & 3.53 & \bf2.8\\
    Interaction\_time(s) & 0.003 & 0.0029 & \bf0.0011\\
    \bottomrule
  \end{tabular}
\end{table}

\section{Discussion and Future Work}
This paper proposes iteration updating and unified representation. Iterative update is used to stabilize the environment and the idea of batch computing is used to save memory and speed up interaction, which largely solves these two problems. This method does not affect decentralized execution and distributed deployment. In addition, our experiments are based on MADDPG, but this method is suitable for most multi-agent algorithms like IQL, VDN, QMIX etc. When combined with PBT or other algorithms for parallel training, the unified representation is particularly efficient.

This method also has some drawbacks. We all know Generative Adversarial Networks are notoriously hard to train, the main difficulty lies in the balance between generator and discriminator and many articles have further studied on it\citep{arjovsky2017wasserstein}\citep{gulrajani2017improved}%\citep{mao2017least}
. In our iterative update, there is also a problem of balancing the capabilities of each agents, especially in mixed cooperative-competitive environments. It is necessary to adjust it cautiously in order to obtain strong agents. At present, we only simply control the learning frequency of iterative update hyperparameter $K$ through experience, which is a research direction in the future. Another problem is how to realize the iterative update method in this unified representative network. The value fixing method based on Bellman Equation can only guarantee a smaller $L_2$ norm of its gradients and cannot strictly hold the equation. This can be further improved in the future work. In addition, due to the limited computing resources, we only expand the number of agents to a certain extent, which can be further verified in more complex environments in the future.

\section{Conclusion}
This paper presents an iterative update and unified representation method to solve the problems of environmental nonstationarity and computational efficiency. This method greatly alleviates the nonstationarity and outperforms MADDPG both in fully-cooperative and mixed cooperative-competitive tasks, especially when the number of agents increases. At the same time, unified representation and batch compute make use of the advantages of tensor compute of neural network, which effectively improves the computing efficiency and avoids the linear growth of the interaction time with the environment. In addition, our method is applicable to most Multi-Agent Reinforcement Learning algorithms. Our future work will focus on the stability of iterative update's training under unified representation and apply this method to more complex tasks.

\clearpage
\small
% \bibliographystyle{named}
% \nocite{*}%添加此处
\bibliographystyle{unsrt}
\bibliography{IUUR}

% % Appendix
% \clearpage
% \normalsize
% \appendix
% \section{Example appendix section}
% \subsection{Example appendix subsection}
% \section{Another example appendix section}
%
% As long as you need contents in appendices.

\end{document}